\documentclass[10pt,twocolumn,letterpaper]{article}

\usepackage{cvpr} 
\usepackage{lineno}
\usepackage{adjustbox} 
\usepackage{array}
\usepackage{caption}
\usepackage{tabularx} 
\usepackage{ulem}

\newcommand\blfootnote[1]{%
  \begingroup
  \renewcommand\thefootnote{}\footnote{#1}%
  \addtocounter{footnote}{-1}%
  \endgroup
}
\definecolor{cvprblue}{rgb}{0.21,0.49,0.74}
\usepackage[pagebackref,breaklinks,colorlinks,allcolors=cvprblue]{hyperref}
\usepackage{graphicx}

\usepackage{booktabs}
\usepackage{multirow}
\usepackage{makecell}
\usepackage{colortbl}

\title{AxisPose: Model-Free Matching-Free Single-Shot 6D Object Pose Estimation via Axis Generation}
\author{
  Yang Zou\textsuperscript{\rm 1}\footnotemark[1], \quad 
  Zhaoshuai Qi\textsuperscript{\rm 1}\footnotemark[1] \footnotemark[2], \quad 
  Yating Liu\textsuperscript{\rm 1}, \quad 
  Zihao Xu\textsuperscript{\rm 2},\\ 
  Weipeng Sun\textsuperscript{\rm 2}, \quad 
  Weiyi Liu\textsuperscript{\rm 1}, \quad 
  Xingyuan Li\textsuperscript{\rm 2}, \quad 
  Jiaqi Yang\textsuperscript{\rm 1}, \quad 
  Yanning Zhang\textsuperscript{\rm 1}\\
  \textsuperscript{1} Northwestern Polytechnical University \hspace{0.1cm}
  \textsuperscript{2} Dalian University of Technology \\
  {\tt\small archerv2@mail.nwpu.edu.cn} \hspace{0.1cm}
  {\tt\small zhaoshuaiqi1206@163.com} \hspace{0.1cm}
}
\begin{document}
\twocolumn[{
\renewcommand\twocolumn[1][]{#1}%
\maketitle

\begin{center}
    \centering
    \captionsetup{type=figure}
    \vspace{-0.15in}
    \includegraphics[width=1\textwidth]{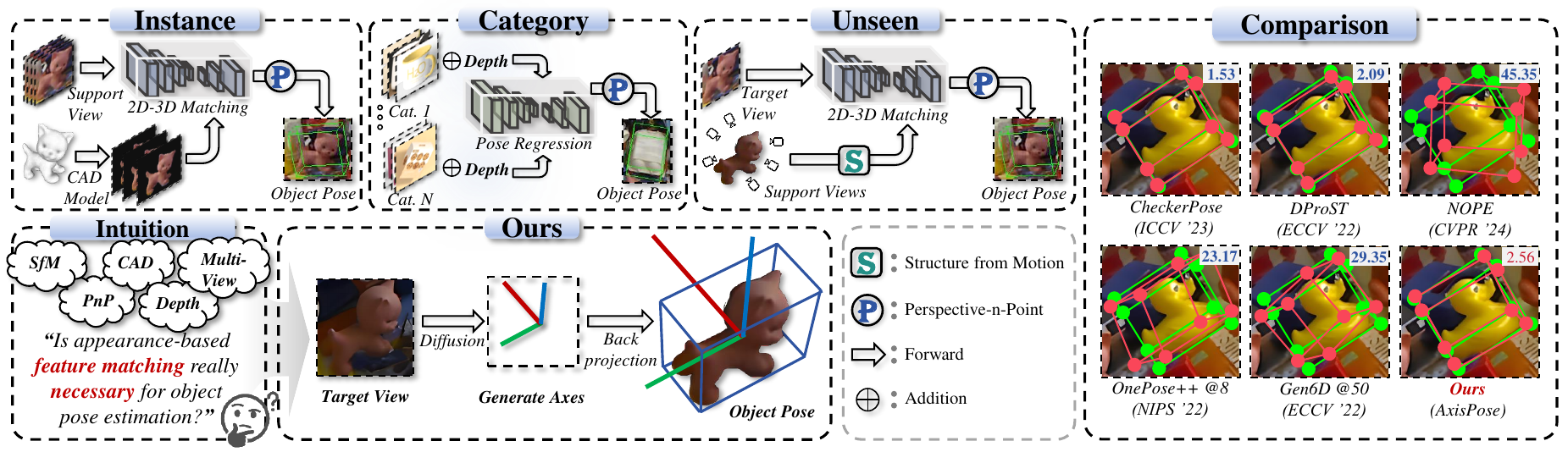}
    \vspace{-0.2in}
    \captionof{figure}{Existing methods rely on direct 2D-3D matching from input CAD models (e.g., instance-level methods) or depth data (e.g., category-level methods) or indirectly from multiple supporting views (e.g., unseen-object methods). In contrast, we hypothesize that each object possesses a tri-axis intrinsic 2D pose representation that reflects its 3D characteristics, making feature matching unnecessary. Based on this idea, we propose inferring the 6D pose in a \textbf{model-free}, \textbf{matching-free}, and \textbf{single-shot} manner by learning the tri-axis as a 2D latent pose representation. We provide a visual comparison with two instance-level methods (CheckerPose~\cite{lian2023checkerpose}, DProST~\cite{park2022dprost}) and three unseen-object methods (NOPE~\cite{nguyen2024nope}, OnePose++\cite{he2022onepose++} with 8 reference views, and Gen6D\cite{liu2022gen6d} with 50 reference views), all retrained in an instance-level manner for fair comparison. The reprojection errors, measured in pixels, are shown in the top right corner.}
    \label{fig:teaser}
\end{center}%

}]
\begin{abstract}
\blfootnote{$^*$ Equal contribution.  $^{\dagger}$ Corresponding author.}
Object pose estimation, which plays a vital role in robotics, augmented reality, and autonomous driving, has been of great interest in computer vision. Existing studies either require multi-stage pose regression or rely on 2D-3D feature matching. Though these approaches have shown promising results, they rely heavily on appearance information, requiring complex input (i.e., multi-view reference input, depth, or CAD models) and intricate pipeline (i.e., feature extraction-SfM-2D to 3D matching-PnP). We propose \textbf{AxisPose}, a \textbf{model-free, matching-free, single-shot} solution for robust 6D pose estimation, which fundamentally diverges from the existing paradigm. Unlike existing methods that rely on 2D-3D or 2D-2D matching using 3D techniques, such as SfM and PnP, AxisPose directly infers a robust 6D pose from a single view by leveraging a diffusion model to learn the latent axis distribution of objects without reference views. Specifically, AxisPose constructs an Axis Generation Module (AGM) to capture the latent geometric distribution of object axes through a diffusion model. The diffusion process is guided by injecting the gradient of geometric consistency loss into the noise estimation to maintain the geometric consistency of the generated tri-axis. With the generated tri-axis projection, AxisPose further adopts a Triaxial Back-projection Module (TBM) to recover the 6D pose from the object tri-axis. The proposed AxisPose achieves robust performance at the cross-instance level (i.e., one model for \(\mathcal{N}\) instances) using only a single view as input without reference images, with great potential for generalization to unseen-object level.
\end{abstract}    
\section{Introduction}
\label{sec:intro}

Object pose estimation is essential for determining the 3D position and orientation of objects in virtual reality (VR), augmented reality (AR), robotics, and 3D scene understanding~\cite{liu2024deep}. Conventional studies mostly explored the instance-level 6D pose estimation problem~\cite{rad2017bb8,tekin2018real,kehl2017ssd}, where the CAD model of the object is available beforehand, limiting its applications in real scenarios. To eliminate the need for CAD models, category-level 6D pose estimation methods are proposed to learn a category-level representation of objects without requiring exact CAD models. These methods estimate the object's pose by learning the intra-category representations, allowing for generalization to new instances within the same category~\cite{ahmadyan2021objectron,wang2019normalized}. However, these methods depend on direct 2D-3D matching with depth, utilizing a complex pose regression network, which restricts their applications when depth data is not available.

Recently, methods for unseen object pose estimation~\cite{sun2022onepose, he2022onepose++, liu2022gen6d, park2020latentfusion} have been proposed to generalize to unseen objects without retraining. OnePose/OnePose++\cite{sun2022onepose, he2022onepose++} matches 2D key points in the query image with 3D points in the SfM model, shifting the focus to 2D-3D feature matching within the established pipeline of feature extraction, SfM, 2D-3D matching, and PnP. Subsequent research efforts have primarily focused on improving the accuracy of 3D representations and feature matching. For instance, the SAM-6D model\cite{lin2024sam} introduces a Sparse-to-Dense Point Transformer to enhance feature matching using the SAM~\cite{kirillov2023segment}. The CF3DGS~\cite{fu2024colmap} reconstructs 3D representations through 3D Gaussian Splatting~\cite{kerbl20233d}. Existing diffusion-based methods, such as 6D-Diff~\cite{xu20246d}, still follow this pipeline, formulating 2D keypoint detection as a reverse diffusion process for better 2D-3D correspondence. Closest to us, NOPE~\cite{nguyen2024nope} estimates the object 3D rotation of the query image from a single reference image via novel-view synthesis but still relies on template matching.

The aforementioned methods fundamentally depend on appearance information from different key points for feature matching. While effective, two major challenges persist: 1) \textbf{Dependence on complex inputs}, such as depth data or at least one reference image, to reconstruct 3D representations (e.g., 3D point clouds~\cite{guo2020deep} and novel view synthesis~\cite{mildenhall2021nerf, kerbl20233d, zou2024enhancing}). These dependencies limit the practicality and scalability of these methods in scenarios where such inputs are unavailable. 2) \textbf{Lack of robustness in degraded environments} arising from heavy reliance on appearance-based matching, which fails in conditions with unreliable visual cues, such as occlusion or weak textures. Given these limitations, we ask, “\textbf{Is appearance-based feature matching really necessary for object pose estimation?}"

The answer is “\textbf{No}." As shown in Figure~\ref{fig:teaser}, we found that the 6D object pose can be directly derived by learning its latent pose representation. We hypothesize that every object possesses an intrinsic 2D pose representation in the form of a tri-axis that resembles its 3D pose characteristics. By learning this 2D latent pose representation, we can infer the 6D pose of the objects. As proved by ~\cite{qi2024indoor}, the 6D pose can be directly derived from an unknown cuboid corner. We then transform the complex problem of object pose estimation into a simplified task of estimating the 2D projections of object axes.

As a response, we propose AxisPose, a model-free, matching-free, single-shot solution for robust 6D pose estimation, which fundamentally diverges from the existing paradigm. Unlike conventional methods that rely on 2D-3D feature matching, our approach generates robust 2D tri-axis projection and, therefore, back-projects to 3D to derive the 6D pose. The key innovation lies in the idea of modeling the latent pose representation through diffusion, which learns the latent distribution of object axes, eliminating the need for appearance-based matching. Specifically, AxisPose proposes an Axis Generation Module (AGM) to capture the latent geometric distribution of object axes through a diffusion model. Also, AxisPose injects the gradient of a designed geometric consistency loss into the noise estimation at each training step, refining the model’s performance across iterations. Inspired by IRUCP~\cite{qi2024indoor}, AxisPose further adopts a Triaxial Back-projection Module (TBM) to recover the 6D pose from the generated 2D projections of object axes. Upon these, AxisPose omits the 3D methods like SfM, PnP, and etc. Our contributions can be summarized as follows:

\begin{itemize}
\item We demonstrate that appearance-based feature matching is not necessary for object pose estimation. Instead, we propose AxisPose, a model-free, matching-free, single-shot solution that models the distribution of latent object axes for robust pose estimation. To the best of our knowledge, this is the first work to approach object pose estimation from a generative perspective.
\item We propose a geometric consistency loss to guide the diffusion process by injecting its gradient into the noise estimation at each training step, progressively refining the model’s performance.
\item We show that the proposed method achieves robust performance at the cross-instance level (i.e., one model for \(\mathcal{N}\) instances) using only a single view as input without reference images, with great potential for generalization to unseen object levels.
\end{itemize}
\section{Related Work}
\label{sec:related}

\subsection{ 6-DoF Object Pose Estimation}
3D input-based methods~\cite{lin2024instance,liu2023istnet,zhang2023genpose,zhang2024lapose} estimate object pose using 3D inputs such as depth data, CAD models, or point clouds. For example, FoundationPose~\cite{wen2024foundationpose} integrates model-based and model-free approaches for multi-task versatility, while IST-Net~\cite{liu2023istnet} learns implicit representations and processes point clouds without explicit shape modeling. DenseFusion~\cite{wang2019densefusion} extracts pixel-wise dense feature embeddings from RGB-D images by processing two data sources individually and then fusing them. Additionally, Normalized Object Coordinate Space (NOCS) shape alignment methods~\cite{tian2020shape,wang2021category,chen2021sgpa} first predict the NOCS shape and then use an offline pose solution to align the object point cloud with the predicted NOCS shape. However, these methods still rely on feature matching and require intricate inputs as priors.

To widen the scope of applications, RGB input-based methods~\cite{he2022onepose++,liu2022gen6d,sun2022onepose,nguyen2024nope,lee2024mfos} have been developed for pose estimation using only RGB images. For example, Gen6D~\cite{liu2022gen6d} performs model-free estimation using a series of reference images, while OnePose and OnePose++\cite{he2022onepose++,sun2022onepose} reconstruct point clouds from RGB images and estimate poses via 2D-3D matching. MFOS\cite{lee2024mfos} leverages a transformer architecture with a set of reference images to estimate unknown object poses, and NOPE~\cite{nguyen2024nope} infers the query image's 3D rotation from a single reference image by estimating a probability distribution over the space of 3D poses. Though effective, these methods fundamentally depend on either keypoint matching or template matching.

\subsection{Diffusion Model}
In recent years, diffusion models have gained significant attention in machine learning, particularly for tasks like image generation, denoising, and translation. Early foundational work in this field was the Denoising Diffusion Probabilistic Model (DDPM)\cite{ho2020denoising}, which uses a Markov process to add noise to data progressively and then learns to reverse this process to generate new samples. ControlNet~\cite{zhang2023adding} injects control conditions into the diffusion process, broadening the application range of diffusion models in image generation. Building on DDPM, Denoising Diffusion Implicit Models (DDIM)\cite{song2020denoising, li2025difiisr} offer a more efficient variant, requiring fewer steps for sample generation while maintaining high quality. Recent advancements, such as Stable Diffusion~\cite{rombach2021highresolution} and Flow-based Diffusion Models~\cite{song2021scorebased}, introduce techniques like multimodal conditioning and flow-based methods to enhance sample quality and diversity.
\begin{figure*}[!htp]
    \centering
    \includegraphics[width=1\linewidth]{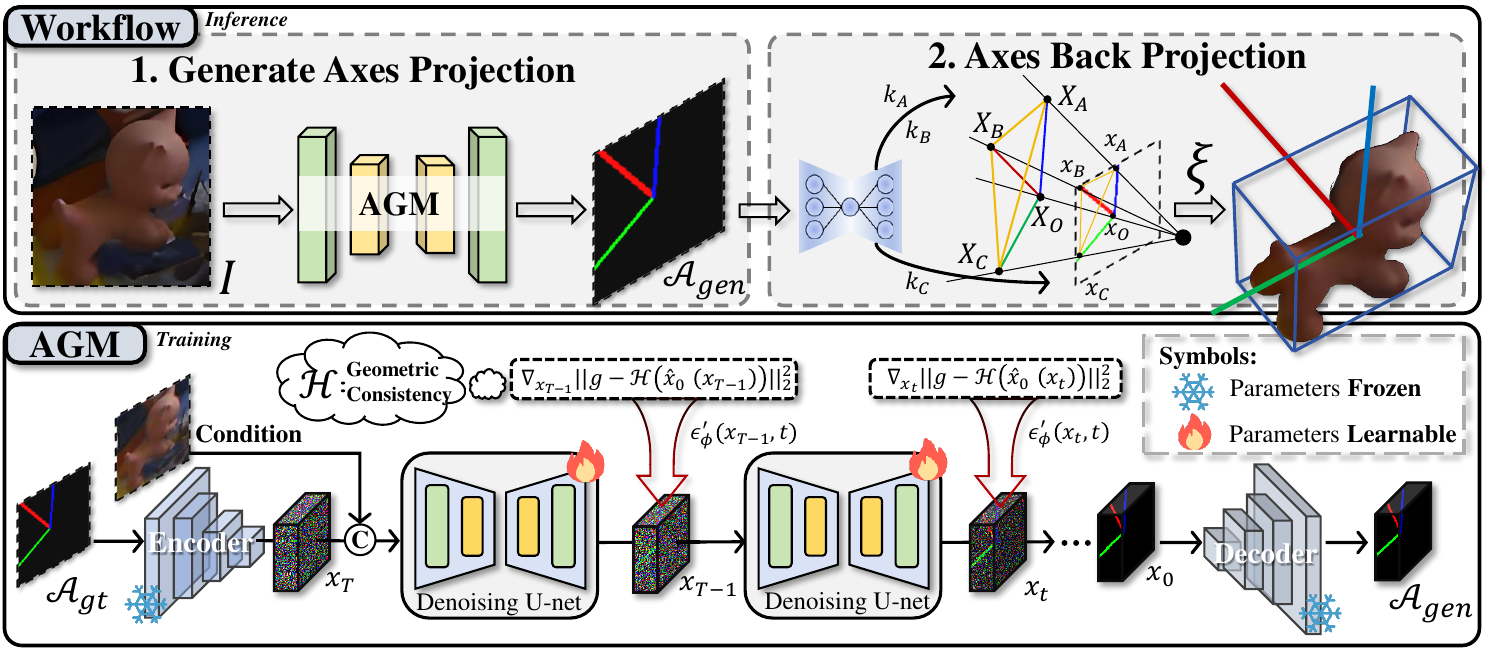}
    \caption{\textbf{Overview of AxisPose.} Given a reference image, the geometric consistency guided Axis Generation Module (AGM) first generates the 2D axes projection. Then, the Triaxial Back-projection Module (TBM) reconstructs the 6D pose from it.}
    \label{fig:overview}
\end{figure*}
\section{Background}

Let \( \mathbf{x}_0 \sim p_{\text {data }}(\mathbf{x}) \) denote samples from the data distribution. Denoising Diffusion Probabilistic Models (DDPMs) iteratively perturb data towards pure noise in a forward process over \( T \) timesteps, applying Gaussian kernels to generate a sequence of latents \( \left\{\mathbf{x}_t\right\}_{t=1}^T \). At each step, noise is added according to a predefined variance schedule \( \left\{\zeta_t\right\}_{t=1}^T \), such that at the final step, the distribution approaches a standard Gaussian, i.e., \(\mathbf{x}_T \sim \mathcal{N}(0, \mathbf{I})\).

Each intermediate latent \( \mathbf{x}_t \) can be directly sampled from a data point \( \mathbf{x}_0 \). The denoising model \( \epsilon_\phi \) is trained to predict the added noise, optimizing the following objective:
\begin{equation}
\mathcal{L}_{\text{simple}}(\phi) = \mathbb{E}_{\mathbf{x}_0, t, \epsilon} \left\| \epsilon_\phi\left( \mathbf{x}_t, t \right) - \epsilon \right\|^2,
\label{eq:simple_loss}
\end{equation}
\noindent where \( t \) is sampled uniformly from \( \left\{1, \dots, T\right\} \), and noise \( \epsilon \) is added to a clean sample \( \mathbf{x}_0 \sim p_{\text {data }} \) to obtain a noisy sample \( \mathbf{x}_t \)~\cite{stathopoulos2024score}.

Denoising Diffusion Implicit Models (DDIMs) extend DDPMs by introducing a non-Markovian sampling process, which allows for more efficient and flexible sample generation. Unlike DDPMs, where the reverse process follows a Markov chain, DDIM directly models the reverse dynamics, enabling faster sampling with the following update rule:
\begin{equation}
\small
\mathbf{x}_{t-1} = \sqrt{\alpha_{t-1}} \hat{\mathbf{x}}_0\left(\mathbf{x}_t\right) 
  + \sqrt{1 - \alpha_{t-1} - \sigma^2} \epsilon_\phi\left(\mathbf{x}_t, t\right) 
  + {n},
\label{eq:update_rule}
\end{equation}
where \( n\sim\mathcal{N}(0,\sigma^{2}\boldsymbol{I}) \), \( \sigma \) is the noise variance during sampling, and \( \hat{\mathbf{x}}_0\left(\mathbf{x}_t\right) \) is the predicted \( \mathbf{x}_0 \), given by:
\begin{equation}
\begin{split}
  \hat{\mathbf{x}}_0\left(\mathbf{x}_t\right) 
    &= \frac{1}{\sqrt{\alpha_t}}\left(\mathbf{x}_t - \sqrt{1 - \alpha_t} \epsilon_\phi\left(\mathbf{x}_t, t\right)\right), \\
    &\simeq \frac{1}{\sqrt{\alpha_t}}\left(\mathbf{x}_t + (1 - \alpha_t) \nabla_{\mathbf{x}_t} \log p\left(\mathbf{x}_t\right)\right).
\end{split}    
\end{equation}

Compared to DDPM, DDIM achieves comparable generation quality while significantly reducing the number of required sampling steps, making it more efficient for practical applications.
\section{Method}
The goal of our approach is to robustly estimate the 6D pose \(\xi\) of an object from a single view. As shown in Figure~\ref{fig:overview}, given a query image \(I\) with its camera intrinsics, we generate the 2D axes projection \(\mathcal{A}_\text{gen}\), where the R, G, and B channels represent the \text{X}, \text{Y}, and \text{Z} axes, respectively. The Axis Generation Module (AGM) follows the DDIM diffusion process to model the latent pose representation, learning the underlying distribution of object axes \(\mathcal{A}_\text{gen}\) while eliminating the need for appearance-based matching. Specifically, AGM is guided by a geometric consistency loss, where the gradient of the guidance loss is injected into the noise estimation at each training step to refine the model progressively. This ensures that the generated axis projections adhere to inherent geometric constraints. Finally, the Triaxial Back-projection Module (TBM) reconstructs the 6D pose from the generated 2D projections of the object axes \(\mathcal{A}_\text{gen}\).

\subsection{Motivation}
The motivation behind our method is quite intuitive. Most existing approaches adhere to the classical pipeline of feature extraction, SfM, 2D-3D matching, and PnP. Fundamentally, these methods rely on appearance-based information from multiple viewpoints to facilitate feature matching, with the core objective being to reconstruct a more accurate 3D representation—whether through depth information or multi-view images. However, what caught our attention is an overlooked aspect of these approaches: after undergoing a complex matching process to estimate 6D object poses, they ultimately validate accuracy by projecting the object’s bounding box into 2D. This raises a crucial question—if the ultimate evaluation is the accuracy of the 2D projection, why not directly predict the 2D projection instead of first estimating the 6D pose? 

Inspired by Qi et al.~\cite{qi2024indoor}, who demonstrated that sufficient constraints on camera-projector intrinsics can be derived from an unknown cuboid corner, we explore a paradigm shift: \textit{Can we directly compute the 2D projection of object poses without explicitly estimating the pose first?} The answer is surprisingly simple—instead of reconstructing the 3D structure for matching, we directly \textbf{generate} the 2D tri-axis projection of the object, treating 3D characteristics as visual features. By leveraging a gradient-injected diffusion model to generate the tri-axis of the object, we demonstrate that 6-DoF pose estimation can be robustly achieved across instances using only a single image, with great potential to extend to unseen instances.

\subsection{Geometric Consistency Guidance}
As shown in Figure~\ref{fig:overview}, since the Axis Generation Module (AGM) generates the 2D projection of the object axes \(\mathcal{A}_\text{gen}\) with DDIM as a baseline, the quality of generation is crucial for the final object pose estimation. Inspired by~\cite{Chung2022ddim}, we introduce a geometric consistency loss as an additional prior and compute its gradient, injecting it into the noise estimation at each step. This guides the diffusion process to generate axis projections that better adhere to inherent geometric constraints. Unlike approaches that bootstrap the inverse process by applying weighted constraints in the final loss function, our method emphasizes posterior sampling. This enables us to make inverse assumptions that reduce dependence on Markov assumptions while preserving the forward inference distribution and accelerating sampling under small step-size constraints.

Specifically, the geometric consistency loss $\mathcal{L}_{\text{geo}}$ consists of two parts: rotation loss \(\mathcal{L}_{\text{rot}}\) and translation loss \(\mathcal{L}_{\text{trans}}\), which can be expressed as:
\begin{equation}
\mathcal{L}_{\text{geo}} = \sum_{i \in \{ \text{X, Y, Z} \}} \overbrace{\left\| \mathcal{A}_{\text{gen}, i} - \mathcal{A}_{\text{gt}, i} \right\|_{2}^{2}}^{\mathcal{L}_{\text{rot}}} + \overbrace{\left\| \mathcal{C}_{\text{gen}} - \mathcal{C}_{\text{gt}} \right\|_{2}^{2}}^{\mathcal{L}_{\text{trans}}},
\end{equation}
\noindent where \(\mathcal{A}_{\text{gen}, i}\) and \(\mathcal{A}_{\text{gt}, i}\) represent the unit vectors for the generated and ground truth axes, respectively, which are derived by normalizing the corresponding axis vectors \(\mathbf{x}_{\text{gen}, i}\) and \(\mathbf{x}_{\text{gt}, i}\) as \(\mathcal{A}_{\text{gen}, i} = \frac{\mathbf{x}_{\text{pred}, i}}{\|\mathbf{x}_{\text{pred}, i}\|_{2}}\) and \(\mathcal{A}_{\text{gt}, i} = \frac{\mathbf{x}_{\text{gt}, i}}{\|\mathbf{x}_{\text{gt}, i}\|_{2}}\). \(\mathcal{C}_{\text{gen}}\) and \(\mathcal{C}_{\text{gt}}\) represent the generated and actual centroid values, respectively.

With the geometric consistency loss, we iteratively incorporate this prior into the inverse diffusion process. This inverse process operates as a denoising procedure, which begins with Gaussian noise and progressively refines the signal through iterative steps, generating structured data that increasingly resembles the original distribution. The noise predicted by the denoising model at time step \( t \) is closely related to the score of the probability density function at the same step~\cite{song2021scorebased}, formulated as:
\begin{equation}
\epsilon_{\phi}(x_{t},t) = -\sqrt{1 - \alpha_{t}} \nabla_{x_{t}} \log p(x_{t}),
\end{equation}
\noindent where \( \nabla_{x_{t}} \log p(x_{t}) \) denotes the score function, i.e., the gradient of the log probability density function with respect to \( x_{t} \).

We define \( g \) as a partial measurement that serves as guidance, obtained by applying the forward operator \( \mathcal{H} \) to the generated RGB triaxial image \( {x}_{0} \). Formally, the forward model is expressed as:
\begin{equation}
g = \mathcal{H}({x}_{0}) + {n}, \quad g, {n} \in \mathbb{R}^{n}, {x}_{0} \in \mathbb{R}^{d},
\label{eq:model}
\end{equation}
\noindent where \( n \) denotes measurement noise, modeled as \( n \sim \mathcal{N}(0, \sigma^{2} \boldsymbol{I}) \). Consequently, the conditional likelihood of \( g \) given \( {x}_{0} \) follows a Gaussian distribution as \( p(g | {x}_{0}) \sim \mathcal{N}(g | \mathcal{H}({x}_{0}), \sigma^{2} \boldsymbol{I}) \). Maximizing the log-likelihood leads to the gradient as \(\nabla\mathcal{L}_{\text{geo}} = \nabla_{x_t}\|g - \mathcal{H}(\hat{x}_{0}(x_{t}))\|_{2}^{2}\).

By leveraging the result \( p(g | x_{t}) \simeq p(g | \hat{x}_{0}) \) from~\cite{Chung2022ddim}, we approximate the gradient of the log-likelihood as: 
\begin{equation}
\nabla_{x_{t}} \log p(g \mid x_{t}) \simeq \nabla_{x_{t}} \log p(g \mid \hat{x}_{0}),
\label{eq:proof}
\end{equation}
\noindent where the latter expression becomes analytically tractable, as the measurement distribution is explicitly defined. Differentiating the likelihood function \( p(g | x_{t}) \) with respect to \( x_{t} \), we obtain:
\begin{equation}
\begin{split}
-\frac{1}{\sigma^{2}} \nabla_{x_{t}} \|g - \mathcal{H}(\hat{x}_{0}(x_{t}))\|_{2}^{2} & \simeq \nabla_{x_{t}} \log p(g \mid \hat{x}_{0}(x_{t})) \\
& \simeq \nabla_{x_{t}} \log p(g \mid x_{t}),
\end{split}
\label{eq}
\end{equation}
\noindent where we explicitly denote \( {\hat{x}}_{0} := {\hat{x}}_{0}(x_{t}) \) to emphasize that \( \hat{x}_{0} \) is a function of \( x_{t} \), and \(\frac{1}{\sigma^{2}}\) is the step size and \(x_{t}\) denotes the initial noisy input.

From this, it follows that \( \nabla_{x_{t}} \log p(g | x_{t}) \) can be expressed using Equation~\ref{eq}, and since \( \nabla_{x_{t}} \log p(x_{t}) \) is known, the score function of the denoising model at time step \( t \), \( \nabla_{x_{t}} \log p(x_{t} | g) \), can be computed via Bayes' theorem. Specifically, we have:
\begin{equation}
\nabla_{x_{t}} \log p(g | x_{t}) = \nabla_{x_{t}} \log p(x_{t} | g) - \nabla_{x_{t}} \log p(x_{t}).
\end{equation}

Using this, we formulate the adjusted noise prediction as:
\begin{equation}
\small
\begin{split}
\epsilon_{\phi}^{\prime} & = -\sqrt{1 - \alpha_{t}} \nabla_{x_{t}} \log p(x_{t} \mid g) \\
& = -\sqrt{1 - \alpha_{t}} [\nabla_{x_{t}} \log p(x_{t}) + \nabla_{x_{t}} \log p(g \mid x_{t})] \\
& = \epsilon_{\phi}(x_{t}, t) + \frac{1}{\sigma^{2}} \sqrt{1 - \alpha_{t}} \nabla_{x_{t}} \|g - \mathcal{H}(\hat{x}_{0}(x_{t}))\|_{2}^{2} \\
& = \epsilon_{\phi}(x_{t}, t) + \frac{1}{\sigma^{2}} \sqrt{1 - \alpha_{t}} \nabla \mathcal{L}_{\text{geo}},
\end{split}
\end{equation}
\noindent where \( \epsilon_{\phi}^{\prime} \) represents the adjusted noise estimate, obtained by incorporating the gradient of the guidance loss \( \mathcal{L}_{\text{geo}} \) into the noise predicted by the original denoising model.

By modifying the noise gradient predicted by the diffusion model, we impose a constraint on the inverse diffusion process, effectively guiding the model towards solutions that better adhere to geometric consistency.

\subsection{Triaxial Back-projection Module}
Inspired by \cite{qi2024indoor}, we propose a Triaxial Back-projection Module (TBM) to recover the 6D pose from a 2D axis projection by back-projecting three orthogonal axes into 3D space. As shown in Figure~\ref{fig:overview}, given a generated 2D axes projection, a regression network estimates the axes’ slopes \( k_\mathrm{A}, k_\mathrm{B}, k_\mathrm{C} \) and the intersection point \( X_O \) in the image plane. These 3D axes \( l_A \), \( l_B \), and \( l_C \) correspond to the edges of a cube corner (C2) defined by vertices \( O, A, B, C \) with image points \( x_O, x_A, x_B, x_C \) and 3D coordinates \( X_O, X_A, X_B, X_C \).

A calibrated camera determines the C2 structure, with only the convex configuration being physically valid. The camera projection model is given by:
\begin{equation}
\lambda
\begin{bmatrix}
x \\
y \\
1
\end{bmatrix}
= 
\begin{bmatrix}
f_{x} & \gamma & x_O \\
0 & f_{y} & y_O \\
0 & 0 & 1
\end{bmatrix}
\begin{bmatrix}
R \mid \lambda_{O} X_{O}
\end{bmatrix}
\begin{bmatrix}
X \\
Y \\
Z \\
1
\end{bmatrix}.
\label{eq:lambda_equation}
\end{equation}

Using the orthogonality of edges \( l_A \), \( l_B \), and \( l_C \) of C2, the following system of equations is derived:
\begin{equation}
\scriptsize
\begin{cases}
\lambda_{A} \lambda_{B} {x}_{A}^{\mathrm{T}} \boldsymbol{\mu} {x}_{B} - \lambda_{A} {x}_{A}^{\mathrm{T}} \boldsymbol{\mu} {x}_{O} - \lambda_{B} {x}_{B}^{\mathrm{T}} \boldsymbol{\mu} {x}_{O} + {x}_{O}^{\mathrm{T}} \boldsymbol{\mu} {x}_{O} = 0 \\
\lambda_{B} \lambda_{C} {x}_{B}^{\mathrm{T}} \boldsymbol{\mu} {x}_{C} - \lambda_{B} {x}_{B}^{\mathrm{T}} \boldsymbol{\mu} {x}_{O} - \lambda_{C} {x}_{C}^{\mathrm{T}} \boldsymbol{\mu} {x}_{O} + {x}_{O}^{\mathrm{T}} \boldsymbol{\mu} {x}_{O} = 0 \\
\lambda_{C} \lambda_{A} {x}_{C}^{\mathrm{T}} \boldsymbol{\mu} {x}_{A} - \lambda_{C} {x}_{C}^{\mathrm{T}} \boldsymbol{\mu} {x}_{O} - \lambda_{A} {x}_{A}^{\mathrm{T}} \boldsymbol{\mu} {x}_{O} + {x}_{O}^{\mathrm{T}} \boldsymbol{\mu} {x}_{O} = 0
\end{cases},
\label{eq:xyz}
\end{equation}

\noindent where \( \lambda_{A}, \lambda_{B}, \lambda_{C} \) are the depth scale factors associated with vertices \( A, B, \) and \( C \). Solving \eqref{eq:xyz} yields the 3D coordinates \( X_O, X_A, X_B, X_C \), from which the 6D pose \(\{ R, T \}\) is derived by aligning the normalized edges:
\begin{equation}
R = \left[\frac{l_A}{\|l_A\|} \quad \frac{l_B}{\|l_B\|} \quad \frac{l_C}{\|l_C\|}\right],
\end{equation}
\noindent with \(\|l_C\| : \|l_B\| : \|l_A\| = r_{C} : r_{B} : 1\), where $r_{B}$ and $r_{C}$ denote the length ratios of legs $l_B$ and $l_C$ with respect to leg $l_A$. The translation is computed as:
\begin{equation}
T = \lambda_{O} X_{O},
\end{equation}

\noindent where \(\lambda_{O}\) is an arbitrary scale factor, which we set to match the scale of the test objects in our experiments.
\section{Experiments}
\subsection{Experimental Settings}
\noindent \textbf{Implementation Details}.  
We implement our method in PyTorch and train it on an NVIDIA A100 GPU. The Axis Generation Module (AGM) and most hyperparameter settings follow the Denoising Diffusion Implicit Models (DDIM) framework \cite{Chung2022ddim}. We incorporate geometric consistency guidance during training to better constrain the generated axes.

\begin{figure*}[!htbp]
	\centering
	\scriptsize
	\renewcommand{\arraystretch}{1.5}  
	\setlength{\tabcolsep}{1.5pt}      
	\begin{tabular}{*{10}{c}}
		\toprule
		\includegraphics[width=0.095\linewidth,height=1.5cm]{ 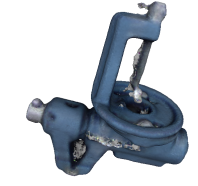} &
		\includegraphics[width=0.095\linewidth,height=1.5cm]{ 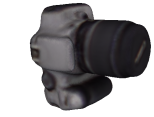} &
		\includegraphics[width=0.095\linewidth,height=1.5cm]{ 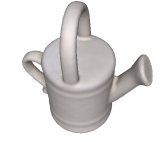} &
		\includegraphics[width=0.095\linewidth,height=1.5cm]{ 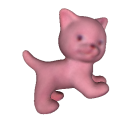} &
		\includegraphics[width=0.095\linewidth,height=1.5cm]{ 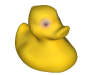} &
		\includegraphics[width=0.095\linewidth,height=1.5cm]{ 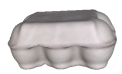} &
		\includegraphics[width=0.095\linewidth,height=1.5cm]{ 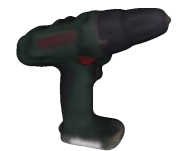} &
		\includegraphics[width=0.095\linewidth,height=1.5cm]{ 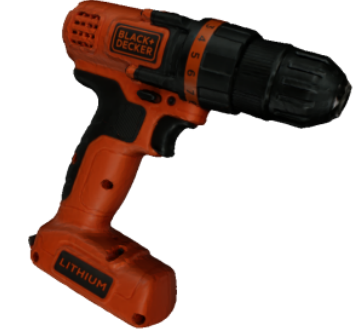} &
		\includegraphics[width=0.095\linewidth,height=1.5cm]{ 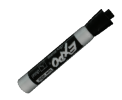} &
		\includegraphics[width=0.095\linewidth,height=1.5cm]{ 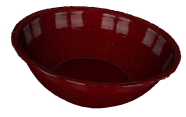} \\
		\textbf{benchvise} & \textbf{cam} & \textbf{can} & \textbf{cat} & \textbf{duck} & 
		\textbf{eggbox} & \textbf{driller} & \textbf{decker} & \textbf{pen} & \textbf{bowl} \\
		\bottomrule
	\end{tabular}
	\caption{\textbf{Visualization of instances used for training and testing.} The first seven instances are from the LINEMOD dataset~\cite{hinterstoisser2012model}, and the remaining three instances come from the YCB-Video dataset~\cite{xiang2017posecnn}. }
	\label{tab:test_objects}
\end{figure*}

\begin{table*}[hbt]
	\caption{\textbf{Quantitative comparison.} “@8” indicates that the method uses 8 reference images. The best results are in \textbf{bold}, while the second-best results are \underline{underlined}. The five model-based instance-level methods are omitted from the ranking for fair comparison, owing to their use of CAD models. }
	\centering
	\renewcommand{\arraystretch}{1}
	\setlength{\tabcolsep}{6pt}
	\begin{tabular}{cccccccccccc}
		\hline
		\multicolumn{1}{c|}{}                                             & \multicolumn{10}{c|}{\textbf{Object Name}}                                                                                                                                                                                   & \multicolumn{1}{l}{}                                \\ \cline{2-11}
		\multicolumn{1}{c|}{}                                             & \textbf{benchvise} & \textbf{cam}   & \textbf{can}   & \textbf{cat}   & \textbf{duck}  & \textbf{eggbox} & \textbf{driller} & \textbf{decker} & \textbf{pen}   & \multicolumn{1}{c|}{\textbf{bowl}}  & \multicolumn{1}{l}{\multirow{-2}{*}{\textbf{Avg.}}} \\ \cline{2-12} 
		\multicolumn{1}{c|}{\multirow{-3}{*}{\textbf{Methods}}}           & \multicolumn{10}{c}{\textbf{Reproj@15pixel}}                                                                                                                                                                                 & \multicolumn{1}{l}{}                                \\ \hline
		\multicolumn{1}{c|}{\textbf{ZebraPose}}                           & 0.965              & 0.932          & 0.873          & 0.989          & 1.000          & 0.448           & 1.000                                    & 0.981           & 0.946          & \multicolumn{1}{c|}{0.931}          & 0.907                                               \\
		\multicolumn{1}{c|}{\textbf{HybridPose}}                          & 0.968              & 0.992          & 0.954          & 0.936          & 0.901          & 0.947           & 0.910                                    & 0.993           & 0.953          & \multicolumn{1}{c|}{0.980}          & 0.953                                               \\
		\multicolumn{1}{c|}{\textbf{DProST}}                              & 0.959              & 0.937          & 0.979          & 0.940          & 0.942          & 0.948           & 0.985                                    & 0.972           & 0.933          & \multicolumn{1}{c|}{0.987}          & 0.958                                               \\
		\multicolumn{1}{c|}{\textbf{GDR-Net}}                             & 0.962              & 0.972          & 0.972          & 0.983          & 0.991          & 0.939           & 0.959                                    & 0.935           & 0.951          & \multicolumn{1}{c|}{0.967}          & 0.963                                               \\
		\multicolumn{1}{c|}{\textbf{CheckerPose}}                         & 0.840              & 0.966          & 0.995          & 0.821          & 0.893          & 0.912           & 0.995                                    & 0.988           & 0.928          & \multicolumn{1}{c|}{0.971}          & 0.931                                               \\ \hline
		\multicolumn{1}{c|}{\cellcolor[HTML]{DEDEDE}\textbf{OnePose++@3}} & 0.192              & 0.219          & 0.153          & 0.365          & 0.477          & 0.308           & 0.153                                    & 0.322           & 0.454          & \multicolumn{1}{c|}{0.080}          & 0.272                                               \\
		\multicolumn{1}{c|}{\cellcolor[HTML]{DEDEDE}\textbf{OnePose++@6}} & 0.192              & 0.406          & 0.347          & 0.344          & 0.477          & 0.625           & 0.643                                    & 0.584           & 0.388          & \multicolumn{1}{c|}{0.454}          & 0.446                                               \\
		\multicolumn{1}{c|}{\cellcolor[HTML]{DEDEDE}\textbf{OnePose++@8}} & 0.575              & 0.500          & 0.373          & 0.557          & 0.556          & 0.754           & 0.683                                    & 0.745           & 0.699          & \multicolumn{1}{c|}{0.842}          & 0.628                                               \\ \hline
		\multicolumn{1}{c|}{\cellcolor[HTML]{DEDEDE}\textbf{Gen6D@50}}    & 0.589              & 0.609          &  0.605    & 0.254          & 0.569          & 0.132           & 0.162                                    & 0.499           & 0.512          & \multicolumn{1}{c|}{0.472}          & 0.440                                               \\
		\multicolumn{1}{c|}{\cellcolor[HTML]{DEDEDE}\textbf{Gen6D@100}}   & 0.645              & {\underline{ 0.908}} & {\underline{ 0.775}}          & 0.534          & 0.667          & 0.593           & 0.522                                    & 0.557           & 0.599          & \multicolumn{1}{c|}{0.534}          & 0.633                                               \\
		\multicolumn{1}{c|}{\cellcolor[HTML]{DEDEDE}\textbf{Gen6D@150}}   & \textbf{0.818}     & \textbf{0.947}    & \textbf{0.966} & \textbf{0.967} & \textbf{0.986} & {\underline{ 0.863}}     & {\underline{ 0.721}}                              & \textbf{0.877}  & {\underline{ 0.901}}    & \multicolumn{1}{c|}{0.849}   & \textbf{0.890}                                      \\ \hline
		\multicolumn{1}{c|}{\cellcolor[HTML]{DEDEDE}\textbf{NOPE}}        & 0.589              & 0.797          & 0.593          & {\underline{ 0.856}}    & 0.587          & 0.799           & 0.369                                    & 0.479           & \textbf{0.976} & \multicolumn{1}{c|}{\textbf{0.931}} & 0.698                                               \\ \hline
		\multicolumn{1}{c|}{\cellcolor[HTML]{DEDEDE}\textbf{Ours}}        & {\underline{ 0.658}}        & 0.719          & 0.661          & 0.785          & {\underline{ 0.788}}    & \textbf{0.868}  & \textbf{0.776}                           & {\underline{ 0.844}}     & 0.847          & \multicolumn{1}{c|}{{\underline{ 0.862}}}          & {\underline{ 0.781}}                                         \\ \hline
		\multicolumn{1}{l}{}                                              & \multicolumn{10}{c|}{\textbf{ADD@0.2d}}                                                                                                                                                                                      &                                                     \\ \hline
		\multicolumn{1}{c|}{\textbf{ZebraPose}}                           & 1.000              & 0.988          & 1.000          & 1.000          & 0.967          & 0.437           & 1.000                                    & 0.967           & 1.000          & \multicolumn{1}{c|}{0.977}          & 0.934                                               \\
		\multicolumn{1}{c|}{\textbf{HybridPose}}                          & 0.996              & 0.959          & 0.936          & 0.979          & 0.803          & 0.996           & 0.870                                    & 1.000           & 0.994          & \multicolumn{1}{c|}{1.000}          & 0.953                                               \\
		\multicolumn{1}{c|}{\textbf{DProST}}                              & 0.998              & 0.985          & 0.996          & 0.973          & 0.875          & 0.997           & 0.991                                    & 0.939           & 0.992          & \multicolumn{1}{c|}{1.000}          & 0.975                                               \\
		\multicolumn{1}{c|}{\textbf{GDR-Net}}                             & 0.962              & 0.993          & 0.943          & 0.982          & 0.985          & 0.973           & 0.826                                    & 0.961           & 0.946          & \multicolumn{1}{c|}{0.972}          & 0.954                                               \\
		\multicolumn{1}{c|}{\textbf{CheckerPose}}                         & 0.810              & 0.922          & 0.957          & 0.623          & 0.699          & 0.700           & 0.937                                    & 0.921           & 0.863          & \multicolumn{1}{c|}{0.899}          & 0.833                                               \\ \hline
		\multicolumn{1}{c|}{\cellcolor[HTML]{DEDEDE}\textbf{OnePose++@3}} & 0.192              & 0.078          & 0.068          & 0.039          & 0.013          & 0.379           & 0.081                                    & 0.042           & 0.129          & \multicolumn{1}{c|}{0.000}          & 0.102                                               \\
		\multicolumn{1}{c|}{\cellcolor[HTML]{DEDEDE}\textbf{OnePose++@6}} & 0.096              & 0.125          & 0.229          & 0.111          & 0.013          & 0.443           & 0.675                                    & 0.411           & 0.117          & \multicolumn{1}{c|}{0.310}          & 0.253                                               \\
		\multicolumn{1}{c|}{\cellcolor[HTML]{DEDEDE}\textbf{OnePose++@8}} & 0.356              & 0.141          & 0.229          & 0.116          & 0.046          & 0.460           & \textbf{0.730}                           & {\underline{ 0.901}}     & 0.141          & \multicolumn{1}{c|}{0.425}          & 0.355                                               \\ \hline
		\multicolumn{1}{c|}{\cellcolor[HTML]{DEDEDE}\textbf{Gen6D@50}}    & 0.534              & 0.422          & 0.522          & 0.149          & 0.397          & 0.126           & 0.153                                    & 0.298           & 0.201          & \multicolumn{1}{c|}{0.386}          & 0.319                                               \\
		\multicolumn{1}{c|}{\cellcolor[HTML]{DEDEDE}\textbf{Gen6D@100}}   & 0.732              & 0.613          & 0.752          & 0.573          & 0.501          & 0.471           & 0.422                                    & 0.392           & 0.389          & \multicolumn{1}{c|}{0.461}          & 0.531                                               \\
		\multicolumn{1}{c|}{\cellcolor[HTML]{DEDEDE}\textbf{Gen6D@150}}   & \textbf{0.918}     & {\underline{ 0.813}}    & \textbf{0.983} & {\underline{ 0.856}}    & {\underline{ 0.768}}    & {\underline{ 0.765}}     & 0.658                                    & 0.712           & 0.834          & \multicolumn{1}{c|}{0.892}          & \textbf{0.820}                                      \\ \hline
		\multicolumn{1}{c|}{\cellcolor[HTML]{DEDEDE}\textbf{NOPE}}        & 0.718              & \textbf{0.875} & 0.241          & \textbf{0.945} & 0.721          & 0.631           & 0.591                                    & 0.623           & \textbf{1.000} & \multicolumn{1}{c|}{{\underline{ 0.989}}}    & 0.733                                               \\ \hline
		\multicolumn{1}{c|}{\cellcolor[HTML]{DEDEDE}\textbf{Ours}}        & {\underline{ 0.877}}        & 0.688          & {\underline{ 0.788}}    & 0.635          & \textbf{0.837} & \textbf{0.772}  & {\underline{ 0.676}}                              & \textbf{1.000}  & {\underline{ 0.871}}    & \multicolumn{1}{c|}{\textbf{1.000}} & {\underline{ 0.814}}   \\ \hline
	\end{tabular}
	\vspace{-3pt} 
	\label{tab_compare}
\end{table*}

\begin{figure*}
	\centering
	\includegraphics[width=0.95\linewidth]{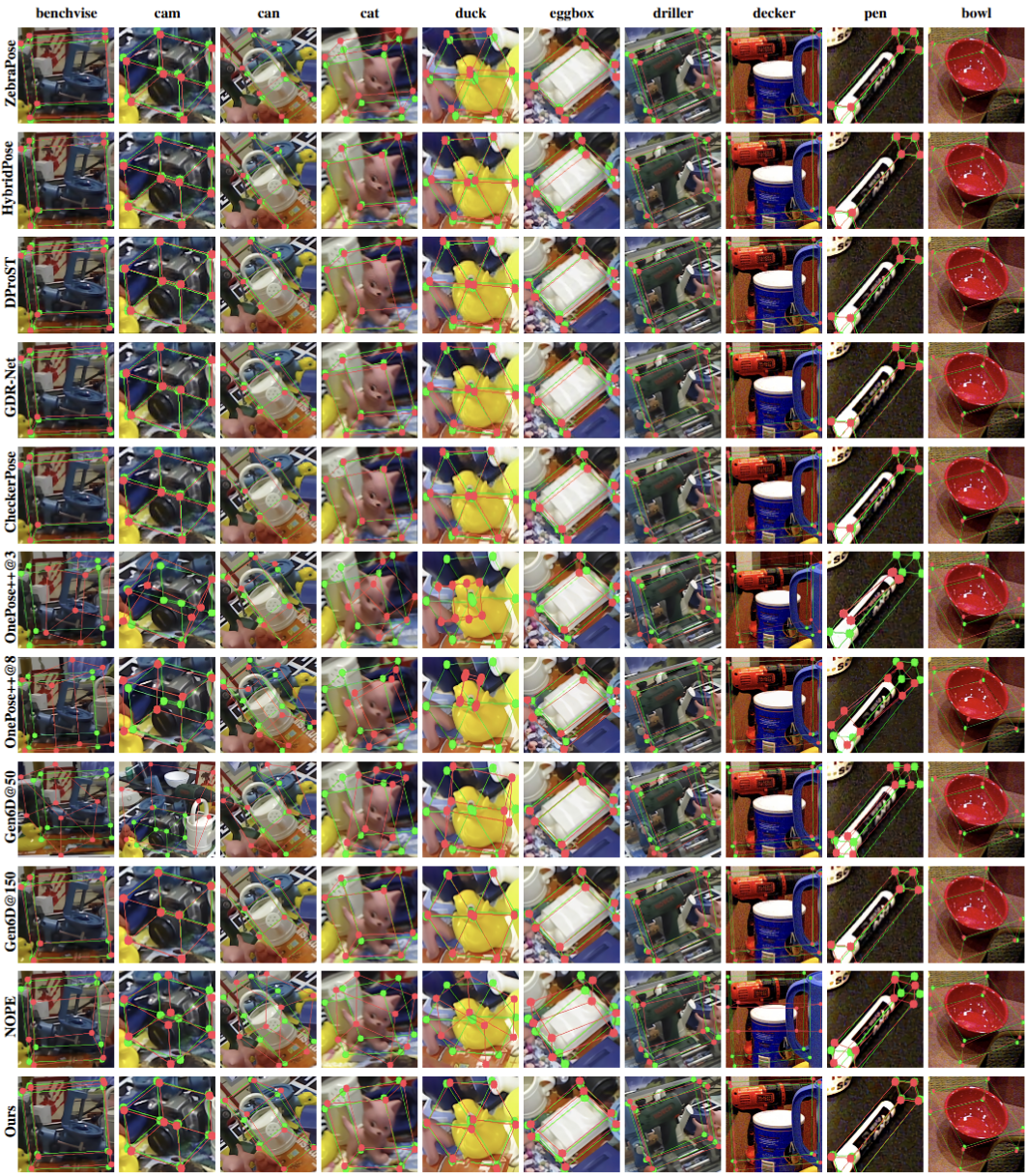}
	\caption{\textbf{Qualitative results.} The green bounding boxes indicate the ground-truth poses, while the red bounding boxes represent the predicted poses. Our method achieves satisfactory performance across various instances and remains robust against degradation of weak texture and occlusion conditions.}
	\label{fig_compare}
\end{figure*}

\noindent \textbf{Datasets and Evaluation Metrics.}
We conduct our experiments on two widely used 6D object pose estimation datasets: LINEMOD (LM)~\cite{hinterstoisser2012model} and YCB-Video (YCB-V)~\cite{xiang2017posecnn}. The LM dataset comprises 13 real-world sequences, each with around 1,200 images of a single object under mild occlusions and cluttered backgrounds. YCB-V is more challenging, containing over 110,000 real images of 21 objects with notable occlusions in cluttered scenes. We train and test our model using a combined dataset of seven objects from LM and three from YCB-V, as illustrated in Figure~\ref{tab:test_objects}. Unlike instance-level methods that train and test a separate model for each instance, our AxisPose trains and tests all instances together in a single model. During training, we apply data augmentation (random cropping, flipping, rotation, and color jittering) to improve robustness.

We evaluate pose estimation using Average Distance Deviation~\cite{hinterstoisser2012model} (ADD@0.2d) and Reprojection Error~\cite{hinterstoisser2012model} at a 15-pixel threshold (Reproj@15pixel). Our core contribution lies in proposing a new solution to object pose estimation in a model-free, matching-free, single-shot manner rather than focusing on state-of-the-art accuracy. Therefore, we adopt a relatively easier threshold for a clearer demonstration. 

\noindent \textbf{Comparative Methods.} We compare our AxisPose with eight state-of-the-art methods, including five model-based instance-level methods (ZebraPose~\cite{su2022zebrapose}, HybridPose~\cite{song2020hybridpose}, DProST~\cite{park2022dprost}, GDR-Net~\cite{Wang_2021_GDRN}, CheckerPose~\cite{lian2023checkerpose}) and three model-free methods (OnePose++~\cite{he2022onepose++}, Gen6D~\cite{liu2022gen6d}, NOPE~\cite{nguyen2024nope}) with varying numbers of reference views. Since NOPE only predicts object rotation, we use our predicted translation for a fair comparison. All model-free methods are retrained at a cross-instance level, consistent with the AxisPose settings.

\subsection{Quantitative Comparison}
Table~\ref{tab_compare} reports the quantitative results. Instance-level methods generally yield high accuracy owing to their use of CAD models. Our AxisPose demonstrates impressive performance for model-free methods compared with OnePose++ and Gen6D under sparse reference views. In particular, AxisPose outperforms OnePose++@3, OnePose++@6, and OnePose++@8 in Reproj@15pixel and achieves higher ADD@0.2d scores than Gen6D@50 and Gen6D@100. Overall, AxisPose ranks first among model-free approaches under minimal input conditions and second among all model-free approaches, indicating that feature matching is not indispensable for robust pose estimation.

\subsection{Qualitative Results}
Figure~\ref{fig_compare} shows the visual qualitative comparison of our AxisPose against the competitors. As expected, model-based methods yield the highest accuracy due to the availability of 3D models. In contrast, model-free approaches often fail under occlusion or in weakly textured instances when only a few reference images are provided. Our AxisPose, however, robustly estimates the pose from a single input image in a model-free, matching-free, single-shot manner, remaining robust even in degraded environments. This outcome further supports the idea that feature matching is not a prerequisite for reliable pose estimation.

\begin{table}
	\begin{minipage}{0.45\textwidth}
		\caption{Quantitative ablation of geometric consistency loss.}
		\centering
		\renewcommand{\arraystretch}{1}
		\setlength{\tabcolsep}{4.5pt}  
		\small
		\begin{tabular}{l|ccccc|c}
			\hline
			\multicolumn{1}{c|}{\textbf{Method}} & \textbf{benchvise} & \textbf{cam}   & \textbf{can}   & \textbf{cat}   & \textbf{duck}  & \multicolumn{1}{l}{\textbf{Avg.}} \\ \hline
			& \multicolumn{5}{c|}{\textbf{Reproj@15pixel}}                                           & \multicolumn{1}{l}{}              \\ \hline
			\textbf{w/o$\sim$$\nabla \mathcal{L}$}         & 0.339              & 0.417          & 0.486          & 0.258          & 0.291          & 0.358                             \\
			\textbf{w$\sim$$\nabla \mathcal{L}$}           & \textbf{0.658}     & \textbf{0.719} & \textbf{0.661} & \textbf{0.785} & \textbf{0.788} & \textbf{0.722}                    \\ \hline
			& \multicolumn{5}{c|}{\textbf{ADD@0.2d}}                                                 &                                   \\ \hline
			\textbf{w/o$\sim$$\nabla \mathcal{L}$}         & 0.421              & 0.526          & 0.366          & 0.313          & 0.325          & 0.390                             \\
			\textbf{w$\sim$$\nabla \mathcal{L}$}           & \textbf{0.877}     & \textbf{0.688} & \textbf{0.788} & \textbf{0.635} & \textbf{0.837} & \textbf{0.765}                    \\ \hline
		\end{tabular}
		\label{table_abla}
	\end{minipage}
\end{table}

\begin{figure}
	\centering
	\includegraphics[width=0.94\linewidth]{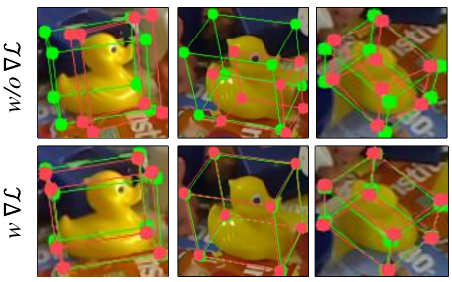}
	\caption{Qualitative ablation of geometric consistency loss.}
	\label{fig_abla}
\end{figure}

\subsection{Ablation Study}
\noindent \textbf{Geometric Consistency Loss.}
We investigate the efficacy of the proposed geometric consistency loss guidance \(\nabla{\mathcal{L}}\) via a comprehensive ablation study. As shown in Table~\ref{table_abla}, removing the geometric consistency loss from AxisPose significantly reduces both the ADD@0.2d and Reproj@15 pixel scores, roughly halving their original values. This underscores the critical importance of this proposed geometric consistency loss guidance for effective pose estimation.

A similar pattern is observed in Figure~\ref{fig_abla}, where omitting the geometric consistency loss leads to markedly poorer pose estimation. Without geometric consistency loss guiding the diffusion process, generated axes become unstable, and their errors are magnified when reprojected into 3D space, leading to a considerable drop in performance.

\begin{figure}
	\centering
	\includegraphics[width=0.98\linewidth]{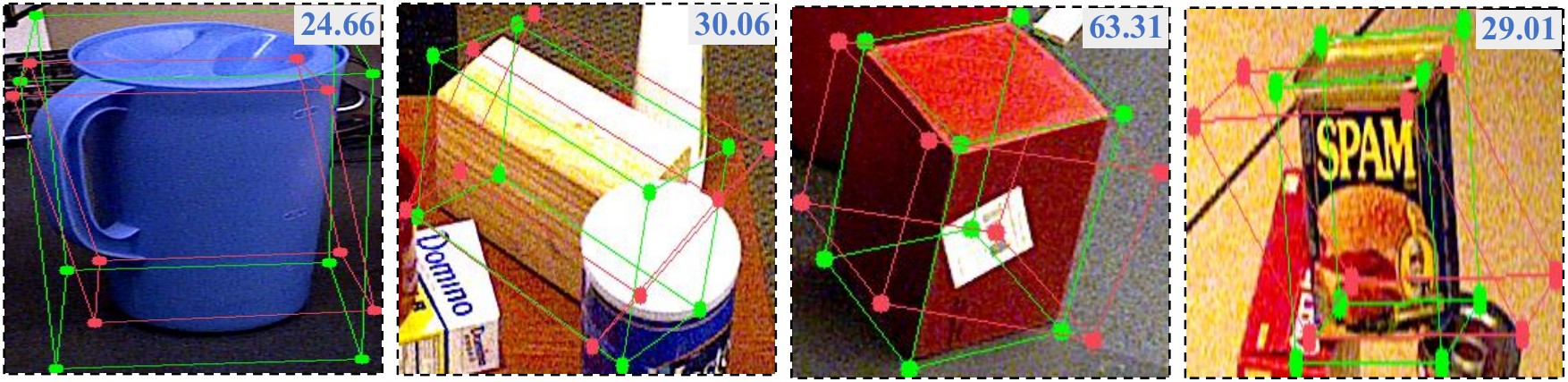}
	\caption{An attempt to extend AxisPose to unseen objects. Further research is needed to achieve robust rotation estimation.}
	\label{fig:unseen}
\end{figure}

\section{Discussion}
We show that appearance-based feature matching is not necessary for robust object pose estimation. Starting from the idea that each object has an intrinsic 2D pose representation that resembles its 3D pose characteristics, we generate the tri-axis projection as the 2D pose representation of objects by diffusion model. Subsequently, we propose a geometric consistency loss to guide the diffusion process by injecting its gradient into the noise estimation at each training step for better axes generation. Finally, we propose a back-projection model to recover the 6D pose from the generated 2D projections of object axes. Upon these, we propose the AxisPose to estimate the object pose in a model-free, matching-free, single-shot manner. Extensive experiments demonstrate the promising performance of the AxisPose.

Unfortunately, its robustness is currently limited at the cross-instance level (one model for \(\mathcal{N}\) instances) and fails to generalize to unseen objects. However, as shown in Figure~\ref{fig:unseen}, AxisPose shows great potential. Our next step is to extend AxisPose to unseen objects while maintaining its model-free, matching-free, single-shot capabilities.
{
    \small
    \bibliographystyle{ieeenat_fullname}
    \bibliography{main}
}

\end{document}